\documentclass[sigconf]{acmart}

\copyrightyear{2027}
\acmYear{2027}
\setcopyright{cc}
\setcctype{by}

\acmConference[WSDM '27]{Proceedings of the Twentieth ACM International Conference on Web Search and Data Mining}{February 15--19, 2027}{Hong Kong, Hong Kong}

\acmBooktitle{Proceedings of the Twentieth ACM International Conference on Web Search and Data Mining (WSDM '27), February 15--19, 2027, Hong Kong, Hong Kong}

\acmDOI{10.1145/3862322.3863872}
\acmISBN{979-8-4007-3117-4/2027/02}

\usepackage{graphicx}

\usepackage{booktabs}
\usepackage{multirow}

\usepackage{amsmath}
\usepackage{bm}

\usepackage{xspace}

\graphicspath{{figures/}}

\begin{document}

% =========================================================
% Title
% =========================================================

\title{What If LLMs Ate Their Words: Causal History Effects in Multi-Turn Interaction}

% =========================================================
% Authors
% =========================================================

\settopmatter{authorsperrow=3}

\author{Jinnan Li}
\orcid{0009-0007-6528-8868}
\email{jnli23@mails.jlu.edu.cn}
\affiliation{%
  \department{School of Artificial Intelligence}
  \institution{Jilin University}
  \city{Changchun}
  \country{China}
}

\author{Zheren Fu}
\orcid{0000-0001-8389-8642}
\email{fzr@ustc.edu.cn}
\affiliation{%
  \institution{School of Cyber Science and Technology, University of Science and Technology of China}
  \city{Hefei}
  \country{China}
}

\author{Yue Wang}
\orcid{0000-0002-0278-2347}
\email{wangyue@email.unc.edu}
\affiliation{%
  \institution{School of Information and Library Science, University of North Carolina at Chapel Hill}
  \city{Chapel Hill}
  \state{North Carolina}
  \country{USA}
}

\author{Jinzhe Li}
\orcid{0009-0002-8975-9920}
\email{jinzhe25@mails.jlu.edu.cn}
\affiliation{%
  \department{School of Artificial Intelligence}
  \institution{Jilin University}
  \city{Changchun}
  \country{China}
}

\author{Yuan Wu}
\orcid{0000-0001-6289-5872}
\authornote{Corresponding authors.}
\email{yuanwu@jlu.edu.cn}
\affiliation{%
  \department{School of Artificial Intelligence}
  \institution{Jilin University}
  \city{Changchun}
  \country{China}
}

\author{Yi Chang}
\orcid{0000-0003-2697-8093}
\authornotemark[1]
\email{yichang@jlu.edu.cn}
\affiliation{%
  \department{School of Artificial Intelligence}
  \institution{Jilin University}
  \city{Changchun}
  \country{China}
}

\renewcommand{\shortauthors}{Li et al.}

% =========================================================
% Abstract
% =========================================================

\begin{abstract}
Multi-turn interaction creates a feedback process in which an LLM's previous responses become context for later behavior.
Prior work shows substantial multi-turn degradation and that assistant-generated history can affect later behavior.
However, it remains unclear how these effects manifest across models, tasks, turns, and inside a model.
% , but the cross-task and cross-model structure, turn-level localization, and internal signatures of these effects remain unclear.
We study these gaps across six task families and five models.
Degradation from fully specified single-turn input (\texttt{FULL}) to progressively revealed multi-turn interaction (\texttt{SHARDED}) is clearly task- and model-dependent,
% exhibits a clear task-by-model structure,
and stronger one-shot performance does not imply greater interaction robustness.
We then retrospectively analyze completed \texttt{SHARDED} conversations by replaying the user messages already observed in each trajectory while editing  only assistant-generated history.
Replacing prior assistant responses with neutral content (termed \emph{neutralization}) changes downstream min-max normalized performance 
by $+.027$ across 2,973 
% paired 
trajectories.
On a prespecified length-controlled subset, short and length-matched neutralization yield nearly identical effects ($+.069$ versus $+.068$).
Turn Surgery further intervenes on one assistant turn at a time.
Among 237 selected degraded trajectories, 63.7\% contain at least one beneficial intervention, while most tested positions remain unchanged; for binary tasks, 48.4\% admit a fail-to-success reversal.
An unedited-replay control on a Qwen3-14B subset further shows that Turn Surgery yields additional gains beyond ordinary regeneration.
An open-weight case study links behaviorally consequential history changes to measurable downstream state differences, but finds task-dependent rather than universal internal signatures.
Overall, assistant-generated history has active but selective effects on multi-turn performance, motivating selective rather than uniform history management.
\footnote{
Code and data are available at:
\url{https://github.com/jinnanli/llms-ate-their-words}
}
\end{abstract}

% =========================================================
% CCS Concepts
% =========================================================



% \begin{CCSXML}
% <ccs2012>
%    <concept>
%        <concept_id>10010147.10010178.10010179.10010181</concept_id>
%        <concept_desc>Computing methodologies~Discourse, dialogue and pragmatics</concept_desc>
%        <concept_significance>500</concept_significance>
%        </concept>
%    <concept>
%        <concept_id>10002944.10011123.10011130</concept_id>
%        <concept_desc>General and reference~Evaluation</concept_desc>
%        <concept_significance>500</concept_significance>
%        </concept>
%  </ccs2012>
% \end{CCSXML}

% \ccsdesc[500]{Computing methodologies~Discourse, dialogue and pragmatics}
% \ccsdesc[500]{General and reference~Evaluation}

% =========================================================
% Keywords
% =========================================================


% =========================================================
% Make Title
% =========================================================

\maketitle

% =========================================================
% Main Paper
% =========================================================

\section{Introduction}
\label{sec:intro}

\begin{figure}[t]
    \centering
    \includegraphics[width=\columnwidth]{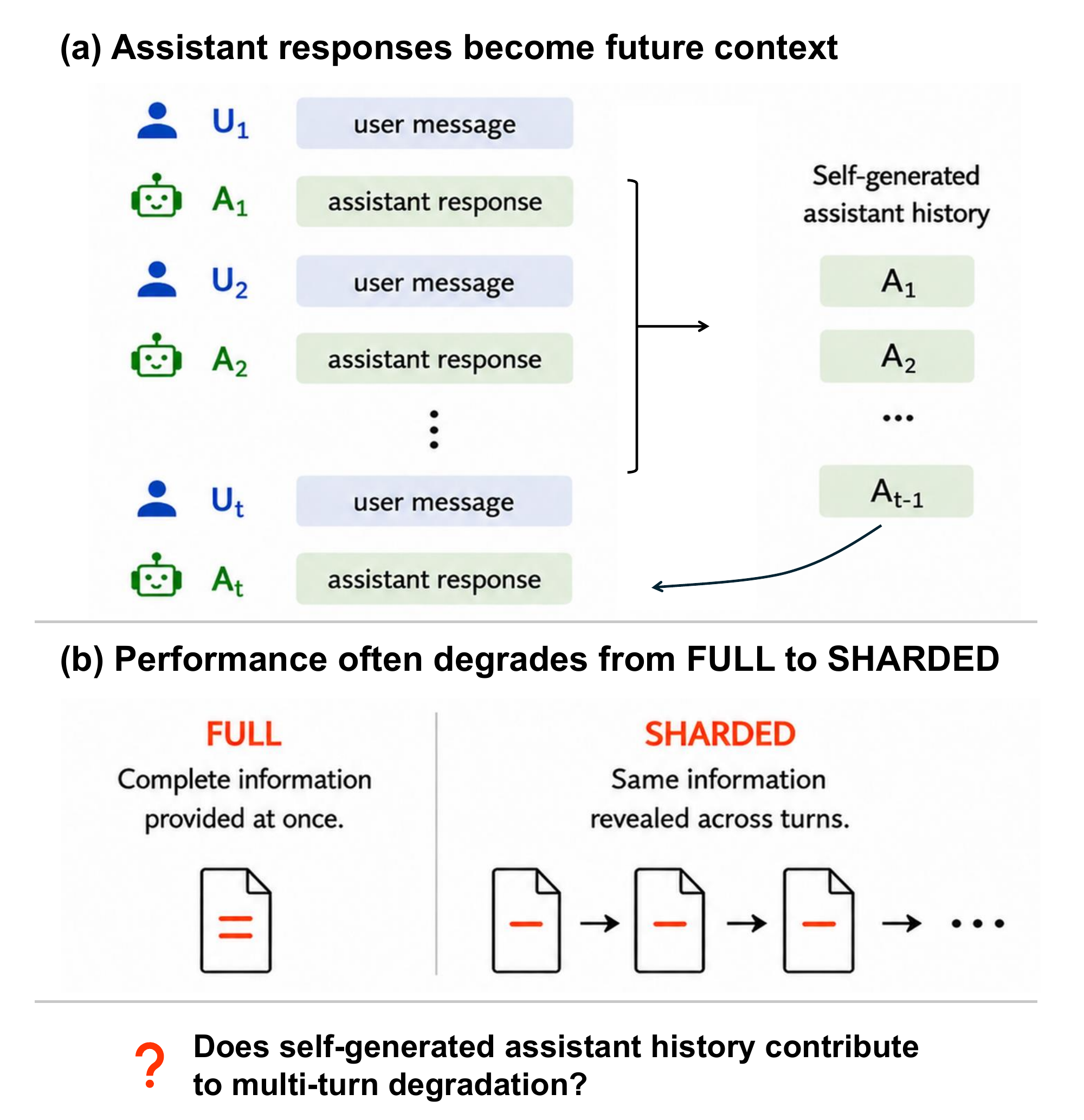}
    \caption{
    Motivation: self-generated assistant history as the future context.
    (a) Assistant responses become part of later context.
    (b) Performance often degrades from \texttt{FULL} to \texttt{SHARDED}, motivating whether this self-generated history contributes to multi-turn degradation.
    }
    \Description{
    Panel (a) shows alternating user messages and assistant responses.
    Arrows illustrate how previous assistant responses are accumulated as self-generated history and fed back into the context for generating subsequent responses.
    Panel (b) contrasts FULL, where complete task information is provided at once, with SHARDED, where the same information is revealed sequentially across multiple turns.
    The question at the bottom asks whether self-generated assistant history contributes to multi-turn degradation.
    }
    \label{fig:motivation}
\end{figure}

Large language models (LLMs) are increasingly used as interactive assistants and agents.
In task-driven settings, users provide requirements over multiple turns, such as revising earlier constraints and continuing the task based on the model's previous responses~\cite{deng2024multi,yao2024tau}.
Unlike single-turn interaction, multi-turn interaction requires models to repeatedly reason over an evolving context, creating new reliability challenges when earlier interaction states influence later decisions.
% Assistant responses are particularly important because they are not only outputs shown to users, but also part of the context for subsequent turns.
We refer to this model-generated and cumulatively retained context as \textbf{self-generated assistant history}.
This raises a natural question: \textit{Is what a model writes into the conversation merely an interaction record, or can it actively influence subsequent task performance?}
Figure~\ref{fig:motivation} illustrates this idea: assistant responses are not only the outputs in the current turn, but also become future input context to influence later behavior.

Prior work introduces the multi-turn degradation as a substantial reliability problem.
\citet{laban2026llms} proposed \emph{sharded simulation}, which compares presenting a complete instruction at once (\texttt{FULL}) with progressively revealing the same information across user--assistant turns (\texttt{SHARDED}).
They show that \texttt{SHARDED} interaction can substantially reduce performance and produce behavioral failures such as premature answers, unsupported assumptions, reliance on previous responses, and loss of intermediate information.
These findings do not isolate whether the model's own previous responses contribute to such failures.
Some other work suggests that a model's own previous responses can become harmful context and bias later behavior~\cite{huang2026llmsbenefitwords,zheng2026maigo,lin2026same}.
However, it remains unclear how these history effects vary across tasks and models, which individual assistant turns are consequential, and whether they are accompanied by measurable changes in internal model states.

We address these gaps through \textit{retrospective analysis of observed conversations}.
Starting from completed conversations, we replay the same user messages while modifying the assistant history.
This allows us to test how previous assistant responses affect later model behavior.
We then compare these effects across tasks and models, identify which assistant turns matter most, and examine whether they are reflected in the internal states of an open-weight model.
Specifically,
\textbf{(RQ1)} we characterize how multi-turn degradation varies across tasks and models.
\textbf{(RQ2)} we test how modifying assistant-generated history changes outcomes, using surface-length and unedited-replay controls.
\textbf{(RQ3)} we introduce \textbf{Turn Surgery}, which modifies one assistant turn at a time and replays recorded user messages to identify influential history turns.
\textbf{(RQ4)} we examine whether behaviorally consequential history interventions have measurable internal-state signatures.
Figure~\ref{fig:rq_overview} summarizes these research questions and their corresponding analyses.
Our main contributions and findings are as follows:

\begin{figure}[t]
    \centering
    \includegraphics[width=\columnwidth]{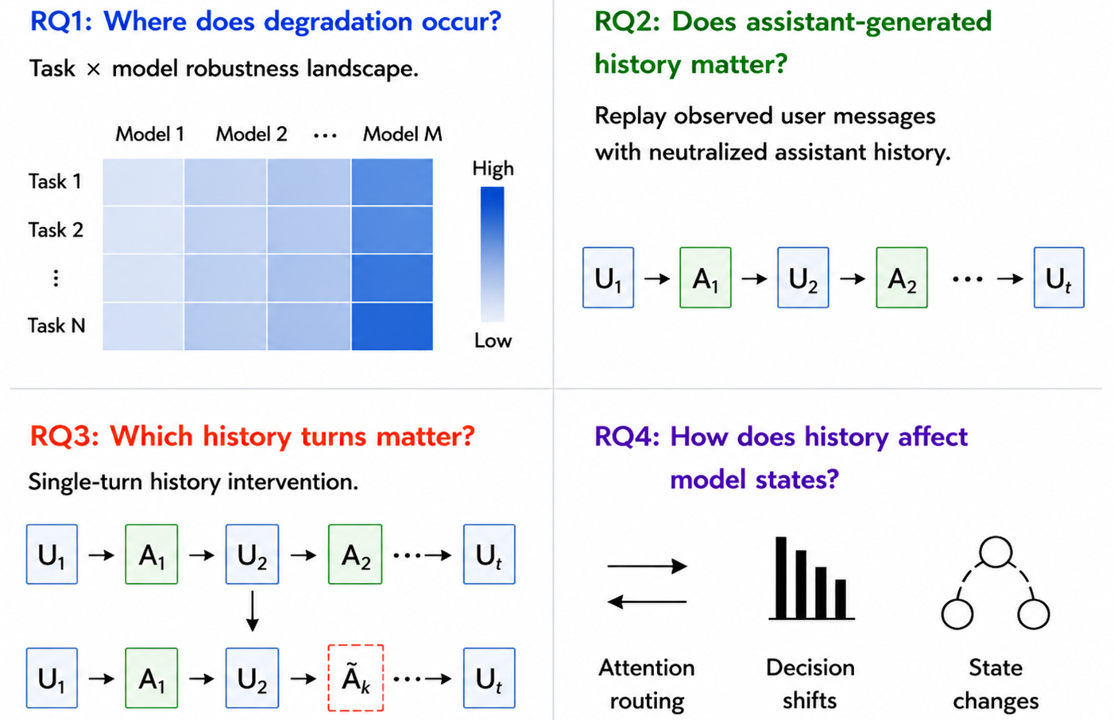}
    \caption{
    Overview of our research work, which covers (1) degradation landscape, (2) assistant-generated history effects, (3) influential turns, and (4) internal model states.
    }
    \Description{
    A four-panel diagram illustrating the four research questions.
    The upper-left panel (RQ1) shows a task-by-model heatmap representing the robustness landscape.
    The upper-right panel (RQ2) depicts a dialogue sequence in which observed user messages are replayed with neutralized assistant history.
    The lower-left panel (RQ3) illustrates a single-turn history intervention by replacing one assistant response while keeping the remaining dialogue context unchanged.
    The lower-right panel (RQ4) depicts three aspects of internal model analysis: attention routing, decision shifts, and state changes, represented by directional arrows, a bar chart, and a node diagram, respectively.
    }
    \label{fig:rq_overview}
\end{figure}

\begin{itemize}
    \setlength{\itemsep}{2pt}
    \setlength{\parskip}{0pt}

    \item \textbf{Multi-turn degradation across tasks and models.}
    Across six task families and five models, \texttt{FULL}--\texttt{SHARDED} degradation exhibits a task $\times$ model structure, and one-shot rankings do not predict interaction robustness.

    \item \textbf{Effects of whole-history neutralization.}
    Across 2,973 paired trajectories, short neutralization produces an aggregate effect of $+.027$.
    Short and whitespace-span-matched neutralization show similar effects, with additional gains confirmed against unedited replay on a Qwen3-14B subset.

    \item \textbf{History effects are turn-selective.}
    Turn Surgery reveals that 63.7\% of selected degraded trajectories contain a beneficial intervention, while most tested positions remain unchanged. 
    An unedited-replay control further confirms a positive mean editing effect on the Qwen3-14B subset.

    \item \textbf{Task-dependent internal signatures.}
    Open-weight analysis reveals measurable downstream state changes associated with influential interventions, with effects varying across tasks.
\end{itemize}

\section{Related Work}
\label{sec:related_work}

\subsection{Multi-turn LLM Evaluation and Underspecified Interaction}
\label{sec:rw_multiturn}

LLM evaluation has increasingly extended from single-turn prompts to interactive settings.
MT-Bench~\cite{DBLP:conf/nips/ZhengC00WZL0LXZ23}, MINT~\cite{DBLP:conf/iclr/00020LCYPJ24}, and subsequent benchmarks such as MT-Eval~\cite{kwan-etal-2024-mt}, MT-Bench-101~\cite{DBLP:conf/acl/BaiLBHLZLSG0O24}, and MultiChallenge~\cite{DBLP:conf/acl/DeshpandeSMJHLK25} extend evaluation toward multi-turn capability, tool use, refinement, and consistency across interaction histories.
Related work on clarification and underspecified interaction further shows that models may answer prematurely or make unsupported assumptions when user intent is incomplete~\cite{pmlr-v244-herlihy24a,luo2025clarifymt}.

The most direct foundation for our study is the sharded simulation framework of Laban et al.~\cite{laban2026llms}.
By progressively revealing shards of the same fully specified task, the framework enables controlled comparison between \texttt{FULL} and \texttt{SHARDED} interaction and demonstrates substantial multi-turn degradation, together with behavioral signatures such as premature answer attempts, unsupported assumptions, and reliance on previous responses.
Recent follow-up work has explored mitigating this degradation through curriculum reinforcement learning, adaptive context resetting, and explicit intent mediation~\cite{li2026mitigating,khalid2025ergo,liu2026intentmismatchcausesllms}.
These studies show that multi-turn degradation can be reduced by changing training or context handling.
We use the same controlled setting to ask a complementary retrospective question: given the user messages already observed in a completed conversation, how much of subsequent behavior is attributable to the model's accumulated assistant history, and do consequential history changes leave measurable internal signatures?

\subsection{Context and Memory in Interactive Foundation Models}
\label{sec:rw_context}

Long-context and conversational-memory research shows that retaining more context does not guarantee that models use it effectively.
Studies of long-context utilization identify strong sensitivity to information position and relevance~\cite{liu2024lost}, while memory systems such as MemGPT~\cite{packer2023memgpt}, LongMemEval~\cite{wu2024longmemeval}, and LoCoMo~\cite{maharana2024evaluating} study how models retain and organize information across longer interactions.
In agentic systems, the context additionally contains model-generated plans, reflections, tool interpretations, and intermediate responses~\cite{shinn2023reflexion}.
Future behavior therefore depends on both externally supplied information and internally generated history.
However, existing studies often evaluate context retention or memory organization at the whole-history level, leaving open how individual assistant-generated states within a trajectory contribute to downstream outcomes.
Related analyses likewise show that earlier model outputs and conversational roles can systematically shape later behavior~\cite{wan2026mitigating,pan2026user}.

\citet{huang2026llmsbenefitwords} study assistant-side history directly by comparing full-context prompting with configurations that omit, summarize, or otherwise reduce previous assistant responses.
They find that assistant history can be unnecessary or harmful and further evaluate history omission with length controls on sharded GSM8K examples derived from Laban et al.~\cite{laban2026llms}.
Other work reduces harmful carry-over across turns through history-cleaned or canonical-context supervision~\cite{zheng2026maigo,lin2026same}, sequential history condensation, or alignment to stronger single-turn anchors~\cite{singh2026mt,chen2026breaking}.

These studies show that managing assistant history can improve multi-turn behavior, largely through whole-history omission or transformation and training-level policies.
We instead ask which assistant turns in observed conversations affect downstream outcomes.
Whole-history neutralization evaluates history editing with a whitespace-span-matched control, while Turn Surgery replaces one intermediate assistant response at a time and replays the recorded user messages to measure position-specific effects.
On a selected Qwen3-14B subset, we further compare edited and unedited replays to distinguish history-editing effects from ordinary regeneration.

\subsection{Behavioral and Counterfactual Analysis of Interaction Trajectories}
\label{sec:rw_trajectory}

A growing line of work analyzes failures at the trajectory level rather than only through terminal task success.
MAST~\cite{NEURIPS2025_b1041e52} and TRAIL~\cite{deshpande2025trail} develop structured taxonomies and annotated traces for diagnosing failures in multi-agent and agentic workflows, while AgentDebug~\cite{zhu2025llm} identifies critical errors and uses targeted feedback to rerun failed trajectories.
These approaches highlight that the step at which a failure becomes visible need not be the step that most strongly determines the final outcome.

More recent work moves from observational diagnosis toward intervention-based attribution.
DoVer~\cite{ma2026dover} validates failure hypotheses by modifying candidate states and rerunning downstream execution.
CausalFlow~\cite{bonagiri2026causalflow} uses step-level counterfactual interventions to identify failure-inducing steps and construct minimal repairs, while Causal Agent Replay~\cite{shah2026causal} estimates how interventions on individual agent steps shift the final outcome distribution.
Turn Surgery shares this intervene-and-replay principle but studies a different object: ordinary assistant turns in controlled multi-turn conversations rather than heterogeneous agent execution traces.
Starting from a completed conversation, it changes one self-generated response and replays the downstream user messages already observed in that trajectory.
This estimates position-specific downstream outcome sensitivity conditional on the recorded user-message history and reveals whether consequential positions are concentrated, distributed, or explained by simple recency.

\section{Method: Evaluation and Intervention Framework}
\label{sec:method}

We first characterize multi-turn degradation using \texttt{FULL} and \texttt{SHARDED}, then use retrospective replay of completed \texttt{SHARDED} trajectories to test whether assistant-generated history affects final outcomes, including a length-matched control.
\textbf{Turn Surgery} applies the same replay logic to one assistant turn at a time to localize turn-level history effects, and RQ4 analyzes internal signatures associated with these interventions in an open-weight model.
Throughout, \emph{causal} and \emph{counterfactual history effects} denote intervention-defined changes conditional on the user messages already observed in the original trajectory and exclude effects mediated through later user adaptation.

\subsection{Sharded Multi-Turn Evaluation Foundation}
\label{sec:method_foundation}

We build on the sharded multi-turn evaluation framework introduced by Laban et al.~\cite{laban2026llms}.
The benchmark contains 627 fully specified instructions across six task families.

For each instruction, we compare two ways of presenting the same task information.
\texttt{FULL} provides the complete instruction in one user message, whereas
\texttt{SHARDED} reveals the information across multiple turns while retaining earlier assistant responses as context.
We write a \texttt{SHARDED} trajectory as $\tau=(U_1,A_1,\ldots,U_T,A_T)$, where $U_t$ and $A_t$ are the user and assistant messages at turn $t$.

The six task families are Actions (105), Code (100), Data-to-Text (120), Database (107), Math (103), and Summary (92).
They are derived respectively from BFCL-V3 Parallel, HumanEval, ToTTo, Spider, GSM8K, and Summary of a Haystack~\cite{patil2025berkeley,chen2021evaluating,parikh2020totto,yu2018spider,cobbe2021training,laban2024summary}.

In the original \texttt{SHARDED} interaction, a user simulator selects and rephrases the next unrevealed shard based on the current conversation state.
Our interventions are retrospective: after a trajectory is generated, we replay its observed user messages rather than rerunning the simulator.
This isolates changes due to assistant history during replay rather than constraining the original user path.
Effects therefore condition on the observed user-message sequence and exclude later user adaptation.

All conditions and interventions are scored on the final assistant output using the benchmark's original task-specific evaluators:
function-call correctness for Actions, test-based pass@1 for Code, SacreBLEU/100 for Data-to-Text~\cite{post2018call}, SQL execution match for Database, normalized exact match for Math, and coverage and citation score for Summary.

\subsection{Model Panel and Statistical Evaluation}
\label{sec:method_models}

We evaluate GPT-5.6 Luna, Grok 4.5, GPT-4o, DeepSeek V3.2, and Qwen3-14B.
All primary generations and replay experiments use temperature $=0$ and a maximum output length of 1,000 tokens.
Qwen3-14B additionally serves as the open-weight model for the internal analysis in RQ4.

The 627 instructions and five models yield 3,135 instruction--model pairs; 2,989 have valid scores under both \texttt{FULL} and \texttt{SHARDED} and enter RQ1; replay analyses use the trace-specific subsets defined below.
Incomplete requests are retained as missing observations rather than converted into failures.
Let $S_i^c$ denote the normalized task-specific evaluator score for case $i$ under condition $c$.

To prevent larger tasks from dominating, model-level summaries macro-average task means with equal weight.
Our degradation measure is $S^{\mathrm{FULL}}-S^{\mathrm{SHARDED}}$; we report model $\times$ task estimates in addition to model-level summaries.
Uncertainty is estimated with 10,000 paired percentile-bootstrap resamples, resampling instructions within task for task-macro statistics.

\subsection{Neutralizing Self-Generated Assistant History}
\label{sec:method_short_ack}

To test whether assistant-generated history affects the final outcome, we retrospectively replay the user messages from each observed \texttt{SHARDED} trajectory while replacing every non-final assistant response with the fixed neutral placeholder $N=\texttt{Acknowledged.}$
This \emph{short neutralization} removes task-specific content while preserving turn structure.
The same model regenerates the final response, which is scored with the original task-specific evaluator.
For trajectory $i$, the history-intervention effect is $S_i^{\mathrm{SHORT}}-S_i^{\mathrm{SHARDED}}$.
Since it also shortens history, Section~\ref{sec:method_length_control} introduces a surface-length control.

Among 3,033 \texttt{SHARDED} trajectories with readable traces and valid scores, 2,973 yield paired short-neutralization outcomes for RQ2.
Our primary estimate averages paired effects within the 30 model $\times$ task cells with equal cell weight, with uncertainty from 10,000 within-cell paired-bootstrap resamples.
We additionally report a trajectory-weighted mean and, for binary tasks, fail-to-success and success-to-failure transitions.

\subsection{Controlling for History Length}
\label{sec:method_length_control}

Short neutralization changes both history content and length.
As a surface-length control, \emph{length-matched neutralization} repeats $N$ to match the surface length of each original non-final assistant response.
Length is measured as the number of non-whitespace spans matched by the regular expression \texttt{\textbackslash S+}; the same estimator and replacement rule are used for all models.
This matches whitespace-delimited spans rather than model tokens.
All 1,250 positive-length replacements in the final controlled subset are matched exactly under this estimator; one additional replacement has zero original length.

From trajectories with valid paired outcomes under \texttt{SHARDED} and short neutralization, we prespecified 10 candidates per model $\times$ task cell (300 total) before observing any length-matched outcomes.
Of these, 264 yield valid outcomes with an actual intermediate-history replacement and form the common paired subset in Figure~\ref{fig:length_control}.
Within this subset, we compare short neutralization with \texttt{SHARDED}, length-matched neutralization with \texttt{SHARDED}, and length-matched with short neutralization.
Overall, task-level, and model-level estimates use model $\times$ task macro aggregation, paired comparisons, and 10,000 percentile-bootstrap resamples.

\subsection{Turn Surgery: Turn-Level Counterfactual Replay}
\label{sec:method_turn_surgery}

Whole-history neutralization cannot identify which earlier assistant responses matter.
\textbf{Turn Surgery} therefore intervenes on one intermediate assistant turn at a time to measure its downstream influence.

\paragraph{Intervention protocol.}
For a target assistant turn $A_{it}$, preceding history is unchanged and $A_{it}$ is replaced with the neutral placeholder $N$.
We replay the recorded downstream user messages while the same model autoregressively regenerates subsequent assistant responses.
The final response is scored with the original evaluator and compared with the recorded \texttt{SHARDED} outcome, yielding the original surgery effect at that history position.

\paragraph{Analysis set.}
Turn Surgery focuses on trajectories with clear \texttt{SHARDED} degradation.
Of 3,135 instruction--model pairs, 3,033 have valid \texttt{SHARDED} traces and 2,989 have complete \texttt{FULL} and \texttt{SHARDED} scores.
For Actions, Code, Database, and Math, a trajectory is eligible when $S^{\mathrm{FULL}}=1$ and $S^{\mathrm{SHARDED}}=0$; for Data-to-Text and Summary, eligibility requires $S^{\mathrm{FULL}}-S^{\mathrm{SHARDED}}\geq0.10$.
This yields 977 replay-eligible trajectories.

Before observing surgery outcomes, we select up to eight per model $\times$ task cell using a fixed deterministic ordering; one cell has only five eligible cases, yielding 237 selected trajectories.
Trajectory-level reach statistics therefore characterize this selected set rather than the full eligible population.
Neutralizing each non-final assistant turn separately produces 1,472 planned branches, of which 1,440 yield valid outcomes and enter the analysis.

\paragraph{Effects and inference.}
For branch $j$ of trajectory $i$, the surgery effect is
$\Delta_{ij}=S_{ij}^{\mathrm{surgery}}-S_i^{\mathrm{SHARDED}}$.
Primary classes use exact sign ($\epsilon=0$): positive if $\Delta_{ij}>0$, no-change if $\Delta_{ij}=0$, and negative if $\Delta_{ij}<0$.
For continuous-score tasks, we additionally assess robustness with $\epsilon=.01$, treating $|\Delta_{ij}|\leq.01$ as no-change and values outside this range by their sign; binary outcomes remain exact.

Because trajectories contain different numbers of valid surgery branches, our primary summary uses a trajectory-equal estimand:
\begin{equation}
\Delta_{\mathrm{traj}}
=
\frac{1}{n_{\mathrm{traj}}}
\sum_{i=1}^{n_{\mathrm{traj}}}
\left(
\frac{1}{k_i}
\sum_{j=1}^{k_i}
\Delta_{ij}
\right),
\end{equation}
where $k_i$ is the number of valid surgery branches for trajectory $i$ and $n_{\mathrm{traj}}$ is the number of trajectories included in the analysis.
Thus, branch effects are first averaged within each trajectory and the resulting trajectory-level means are then weighted equally.
We additionally report a branch-weighted mean as a robustness check.
Turn-level outcome composition follows the same trajectory-equal principle: branch proportions are computed within each trajectory and then averaged equally across trajectories.
Trajectory-level reach records whether a trajectory contains at least one positive surgery.

For binary tasks, a fail-to-success branch changes the final score from 0 under \texttt{SHARDED} to 1 after surgery.
We report the trajectory-level proportion containing at least one such reversal; this analysis covers 157 binary-task trajectories.

We additionally characterize where positive surgery effects occur and how many such positions appear within each trajectory.
For positional analysis, relative position denotes the normalized location of the intervened assistant turn, while distance-to-final measures the number of subsequent user turns between that turn and the final user request.
These branch effects form a turn-level sensitivity signal over historical positions.

Inference uses 10,000 trajectory-clustered bootstrap resamples, each retaining all valid surgery branches from a sampled trajectory.

We additionally assess repeatability on a frozen subset of 60 trajectories.
For each, we select the intervention with the largest first-run effect magnitude and rerun it twice under the same temperature-$0$ configuration.
Exact outcome-class consistency requires all three runs to share the same positive, no-change, or negative class; effect-sign agreement is the proportion of the three run pairs in the same class, treating no-change separately.
This stability analysis concerns selected high-magnitude interventions rather than all Turn Surgery branches.

\subsection{Unedited-Replay Controls}
\label{sec:method_null_replay}

To distinguish history editing from ordinary regeneration, we introduce an unedited-replay control on 40 selected Qwen3-14B trajectories across five tasks.

For Turn Surgery, the unedited branch preserves the original target assistant response and regenerates all subsequent assistant turns under the recorded user messages. The edited branch replaces that response with $N$ and regenerates the same downstream turns. We measure the net editing effect as
\begin{equation}
\Delta^{\mathrm{net}}_{ij}
= S_{ij}^{\mathrm{surgery,new}}
- S_{ij}^{\mathrm{null,new}}.
\end{equation}
This analysis covers 219 surgery branches.

For whole-history neutralization, we use the same 40 trajectories and regenerate only the final response under three conditions: unchanged history, short neutralization, and whitespace-span-matched neutralization. Both analyses use paired comparisons and 10,000 trajectory-level bootstrap resamples, with surgery effects averaged within each trajectory.

\subsection{Open-Weight Internal Analysis}
\label{sec:method_internal_analysis}

To examine how behaviorally consequential history interventions are reflected inside the model, we conduct an open-weight case study on Qwen3-14B.
RQ4 asks whether positive-effect surgeries produce larger downstream state changes, whether restoring those states toward the original-history state can partially reverse the behavioral effect, and whether complementary probes show a consistent pattern.
The primary internal analyses cover Actions, Code, and Data-to-Text, the three task families for which exact benchmark evaluation of newly generated outputs is locally available.
Probe definitions were frozen on a 24-trajectory discovery set spanning these tasks; primary estimates use a disjoint held-out set of 118 trajectories and 603 surgery branches, including 46 trajectories with both positive and no-change interventions.
Branch labels are recomputed from Qwen3-14B task-score deltas under the same Turn Surgery protocol; primary held-out probe contrasts compare positive with no-change branches.
A separate prospective Data-to-Text (D2T) population is used only for replication and is not pooled with the held-out set.

\paragraph{State propagation.}
We first ask whether positive-effect surgeries produce larger changes in downstream model states.
At each decoder layer, we extract the residual stream, the representation passed between transformer blocks, at the final content token of each replayed downstream user message.
These user messages are already part of the observed trajectory and are replayed unchanged while the assistant-history intervention varies.
We call these positions \emph{downstream user anchors} and the last one the \emph{final downstream anchor}.
Original-versus-surgery representation divergence is measured as one minus cosine similarity and summarized both at the final anchor and by a normalized trapezoidal area over all downstream anchors (\emph{propagation AUC}).
Because positive and no-change branches differ in task composition, we report pooled, equal-task-macro, and within-trajectory contrasts, with stricter positional controls using nearest relative position or distance-to-final difference at most one.

\paragraph{State restoration.}
We next ask whether these downstream state changes are directionally related to the behavioral effect.
At the final downstream anchor, we perform residual-state patching that moves a surgery context back toward its corresponding original-history state.
At each tested layer, the surgery-context residual vector is replaced once during prefill with the corresponding original-history vector, after which deterministic greedy generation proceeds normally.
These \emph{restoration patches} measure how much of the surgery benefit is undone by restoring this downstream state.
The held-out cohort uses one positive--no-change pair from each of 46 mixed trajectories; Figure~\ref{fig:internal_signatures}B reports the 33 D2T pairs alongside 43 prospective D2T pairs.
D2T restoration is evaluated over predefined layer ranges 3--12 and 16--31; the cohorts are analyzed separately, and patching is treated as a bounded causal intervention rather than a mediation test.

\paragraph{Additional internal probes.}
Finally, we ask whether simpler internal signals provide a consistent signature of behavioral influence.
We measure static target-turn attention and original-to-surgery redistribution of aggregate attention to prior assistant history.
For attention knockout, we block the target assistant turn either only from final generated tokens (\emph{direct-only}) or from all downstream query positions (\emph{full-downstream}); their difference in absolute effect is the \emph{propagation advantage}.
We also measure the \emph{continuation margin}, the difference between length-normalized teacher-forced log-likelihoods of successful and failed continuations, and examine how it changes between original and surgery contexts.
These probes are treated as complementary diagnostics of task dependence rather than evidence for a single universal mechanism.

\section{Results}
\label{sec:results}

We present results for multi-turn degradation (RQ1), whole-history intervention (RQ2), turn-level sensitivity (RQ3), and internal model signatures (RQ4).

\subsection{RQ1: Multi-turn Degradation Forms a Structured Task $\times$ Model Landscape}
\label{sec:results_rq1}

\begin{figure*}[t]
    \centering
    \includegraphics[width=\textwidth]{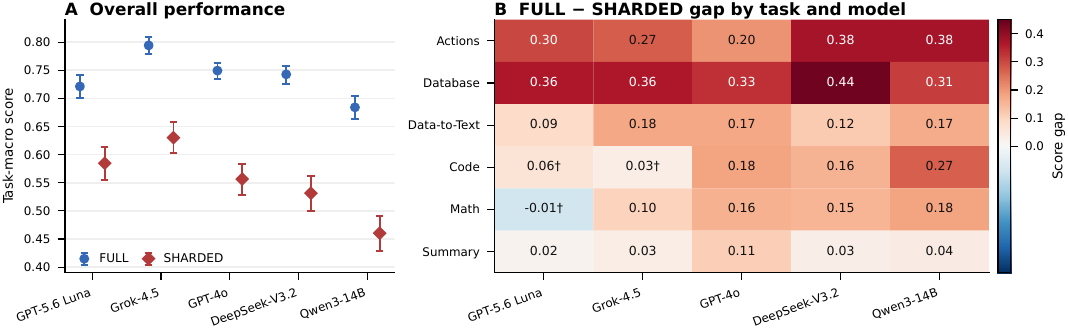}
    \caption{
    Multi-turn degradation exhibits a structured task $\times$ model landscape.
    (A) Overall task-macro performance under \texttt{FULL} and \texttt{SHARDED}, showing the aggregate degradation from one-shot input to multi-turn interaction.
    (B) The corresponding \texttt{FULL}-\texttt{SHARDED} gaps across tasks and models, revealing heterogeneous degradation patterns.
    Error bars denote 95\% paired-bootstrap confidence intervals; $\dagger$ indicates intervals including zero.
    }
    \Description{
    Two panels compare multi-turn performance across five models and six tasks.
    Panel A is a point plot with confidence intervals, where blue circles represent FULL and red diamonds represent SHARDED.
    All five models achieve lower task-macro scores under SHARDED.
    Panel B is a heatmap with tasks as rows and models as columns.
    Each cell displays the numerical score gap between FULL and SHARDED, with red indicating positive gaps and blue indicating negative gaps.
    Database and Actions exhibit consistently large degradation across models, while Summary exhibits relatively small gaps.
    Code and Math show greater variation across models.
    The largest gap is 0.44 for DeepSeek-V3.2 on Database, and the smallest is -0.01 for GPT-5.6 Luna on Math.
    Dagger symbols indicate confidence intervals that include zero.
    }
    \label{fig:robustness}
\end{figure*}

All five models degrade when moving from fully specified input to progressively sharded interaction, but the magnitude of this loss varies substantially across models and tasks.
As shown in Figure~\ref{fig:robustness}A, the absolute \texttt{FULL}--\texttt{SHARDED} gaps are .137 for Luna, .164 for Grok, .193 for GPT-4o, .211 for DeepSeek, and .224 for Qwen.

Notably, model rankings under \texttt{FULL} do not match rankings by \texttt{FULL}--\texttt{SHARDED} degradation.
Grok achieves the highest \texttt{FULL} score but does not have the smallest gap, whereas Luna does not lead under \texttt{FULL} yet exhibits the smallest absolute gap.
Although absolute gaps may still depend partly on starting performance, stronger one-shot performance therefore does not necessarily imply greater multi-turn robustness.

Figure~\ref{fig:robustness}B further shows that degradation follows a structured task $\times$ model pattern.
Actions and Database exhibit relatively large \texttt{FULL}--\texttt{SHARDED} losses across models.
Data-to-Text shows moderate degradation with comparatively consistent behavior across models, whereas Code and Math exhibit stronger model dependence.
Summary is the most stable overall.
Thus, multi-turn degradation is not a uniform property of a model but depends systematically on the task and model combination.

Together, RQ1 shows that multi-turn robustness is distinct from one-shot capability and exhibits substantial task $\times$ model structure.
Because \texttt{SHARDED} interaction also requires models to repeatedly condition on assistant responses they previously wrote into the conversation, RQ2 asks whether this self-generated history is itself a causal contributor to downstream performance.

\subsection{RQ2: Self-Generated Assistant History Shapes Downstream Performance}
\label{sec:results_rq2}

\subsubsection{Large-Scale History Intervention Changes Downstream Outcomes}

Assistant-history neutralization changes downstream outcomes on average under retrospective replay of the recorded user-message sequence.
Among 3,033 \texttt{SHARDED} trajectories with readable traces and valid benchmark scores, short neutralization yields 2,973 paired outcomes.
Across this population, the equal-cell model $\times$ task macro score is .550 under \texttt{SHARDED} and .577 under short neutralization, corresponding to an overall effect of $+.027$ with a 95\% CI of $[.013,.041]$.
Assigning equal weight to paired trajectories yields a similar effect of $+.033$ with a 95\% CI of $[.018,.048]$.

This average masks substantial heterogeneity across models and tasks.
Luna, Grok, GPT-4o, and Qwen show positive effects under short neutralization, whereas DeepSeek exhibits a clear negative effect of $-.149$ $[-.186,-.111]$.
At the task level, Actions shows a large positive effect of $+.223$ $[.181,.265]$, whereas Code shows a clear negative effect of $-.133$ $[-.180,-.087]$.
Database has a point estimate of $+.036$ $[-.007,.080]$, while the remaining tasks show smaller positive, near-zero, or mildly negative effects.
Overall, these results show that assistant-generated history has measurable downstream effects, but the direction and magnitude of those effects vary substantially across models and tasks.
Across the 30 model $\times$ task cells, multi-turn degradation and short-neutralization effects show a positive descriptive association (Spearman $\rho=.553$).
However, Actions benefits from neutralization across all five models, while Code shows negative effects in four models and Database exhibits mixed directions.

Because short neutralization changes both history content and length, we next test whether this history effect can instead be explained by shorter surface history.

\subsubsection{History Effects Under Surface-Length Matching}

On the prespecified subset, the neutralization effect persists under whitespace-span matching.
From trajectories paired under \texttt{SHARDED} and short neutralization, we prespecified 300 length-control candidates using model $\times$ task stratification.
Of these, 264 yield valid outcomes with an actual intermediate-history replacement and form the common paired subset for all three conditions.
As described in Section~\ref{sec:method_length_control}, every positive-length replacement is matched exactly under the shared whitespace-span estimator.

\begin{figure}[t]
    \centering
    \includegraphics[width=\columnwidth]{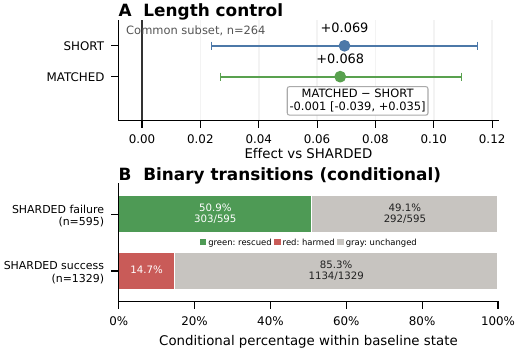}
    \caption{
    History effects persist under surface-length matching.
    (A) Effects of short and length-matched neutralization relative to \texttt{SHARDED} on the common subset of 264 trajectories.
    (B) Conditional binary outcome transitions under short neutralization among trajectories that initially succeed or fail under \texttt{SHARDED}.
    Error bars denote 95\% paired-bootstrap confidence intervals.
    }
    \Description{
    Two vertically arranged panels illustrate the effects of assistant history neutralization.
    Panel A shows a horizontal point plot with confidence intervals.
    Short and length-matched neutralization yield nearly identical improvements of 0.069 and 0.068 relative to SHARDED, respectively. Their difference is -0.001, with a 95\% confidence interval from -0.039 to 0.035.
    Panel B shows two horizontal stacked percentage bars, with green indicating rescued trajectories, red indicating harmed trajectories, and gray indicating unchanged outcomes.
    Among 595 initially failed trajectories, 50.9\% are rescued and 49.1\% remain unsuccessful.
    Among 1,329 initially successful trajectories, 14.7\% become unsuccessful and 85.3\% remain successful.
    }
    \label{fig:length_control}
\end{figure}

On this common subset, Figure~\ref{fig:length_control}A shows effects of $+.069$ $[.024,.115]$ for short neutralization and $+.068$ $[.027,.110]$ for length-matched neutralization, both relative to \texttt{SHARDED}.
The difference between the two neutralization conditions is only $-.001$ $[-.039,.035]$.
Because the full-population and controlled estimates use different populations, the length-control inference rests on this paired 264-trajectory comparison.

The controlled task-level results remain heterogeneous.
Under length-matched neutralization, the effect is $+.260$ $[.160,.380]$ for Actions and $+.140$ $[.020,.280]$ for Database, whereas Data-to-Text shows a small negative effect of $-.022$ $[-.061,.017]$.
A supplementary tokenization audit on available Qwen3-14B histories reveals substantial differences in actual prompt-token counts under whitespace-span matching.

\paragraph{Unedited-replay control.}
On 40 selected Qwen3-14B trajectories across five tasks, we compare short and whitespace-span-matched neutralization with unedited final-answer replay.
Relative to unedited replay, short and matched neutralization improve the mean score by $+.358$ (95\% CI $[.215,.510]$) and $+.367$ ($[.223,.518]$), respectively.
Their paired difference is $+.010$ ($[-.069,.088]$).
The largest gains occur in Actions, where all eight selected degraded trajectories improve under both interventions.

\subsubsection{History Intervention Produces Heterogeneous Outcome Transitions}

The positive aggregate effect also masks bidirectional trajectory-level changes.
Figure~\ref{fig:length_control}B examines binary outcomes under short neutralization across the retrospective replay population.
Among 1,924 binary-task trajectories, 303 of 595 baseline failures are rescued (50.9\%), whereas 195 of 1,329 baseline successes are harmed (14.7\%).

The same uniform history intervention can therefore improve some trajectories while degrading others.
This trajectory-level heterogeneity motivates a more selective question: which specific history positions help or hurt downstream behavior?
RQ3 addresses this by intervening on individual assistant turns.

\subsection{RQ3: Turn Surgery Reveals Broad but Structured History Sensitivity}
\label{sec:results_rq3}

\begin{figure*}[t]
    \centering
    \includegraphics[width=\textwidth]{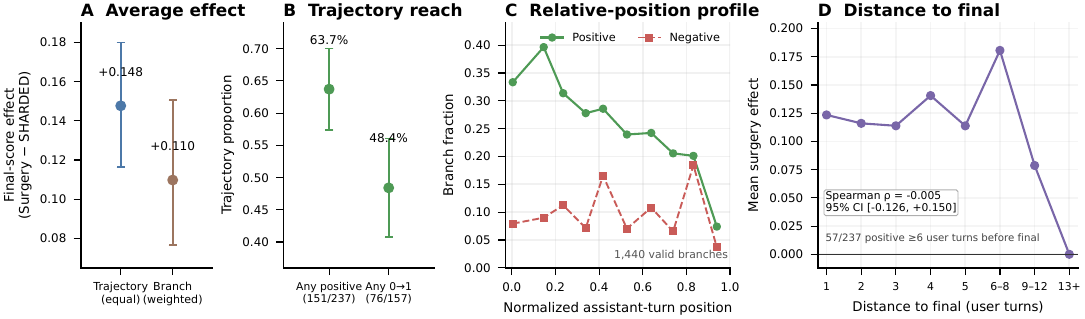}
    \caption{
    Turn Surgery reveals broad but structured history sensitivity within the selected degraded trajectories.
    (A) Average surgery effects under two aggregation schemes.
    (B) Trajectory-level reach of positive interventions and fail-to-success reversals.
    (C) Turn-level positive and negative effects across normalized assistant-turn positions, where 0 and 1 correspond to the earliest and latest intermediate assistant turns within each trajectory, respectively.
    (D) Surgery effects across distance-to-final bins.
    }
    \Description{
    Four panels summarize the structure of turn-level history sensitivity under Turn Surgery.
    Panel A shows average surgery effects under two aggregation schemes, with a trajectory-level equal average effect of +0.148 and a branch-weighted average effect of +0.110, both with confidence intervals.
    Panel B shows trajectory-level reach, indicating that 63.7\% of trajectories have at least one positive intervention and 48.4\% have at least one fail-to-success reversal.
    Panel C plots positive and negative branch fractions across normalized assistant-turn positions, with positive effects generally more common than negative effects across most positions.
    Panel D plots mean surgery effect by distance to the final turn, showing variation across distance bins rather than a simple monotonic recency pattern.
    The far-right bin labeled 13+ has an effect near zero, and the panel also reports a Spearman correlation near zero with a confidence interval spanning zero.
    }
    \label{fig:turn_surgery}
\end{figure*}

Turn Surgery reveals that history sensitivity is broad across degraded trajectories but selective at the level of individual assistant turns.
As described in Section~\ref{sec:method_turn_surgery}, from 977 replay-eligible degraded trajectories, we select 237 model $\times$ task-stratified trajectories before observing surgery outcomes and neutralize each intermediate assistant turn separately.
All trajectory-level proportions below characterize this selected analysis set.
Figure~\ref{fig:turn_surgery} summarizes the average effects, trajectory-level reach, turn-level selectivity, and positional structure of these interventions.

\subsubsection{Average Effect and Trajectory Reach}

Within the selected degraded set, Turn Surgery has a positive average effect, and positive interventions occur across many trajectories rather than being driven by a small number of cases.
Using the trajectory-equal estimand, Figure~\ref{fig:turn_surgery}A shows a mean surgery effect of $+.148$, with a 95\% CI of $[.116,.180]$ relative to the original \texttt{SHARDED} trajectory.
Assigning equal weight directly to all surgery branches yields a branch-weighted effect of $+.110$ with a 95\% CI of $[.077,.151]$, showing that the positive average is robust to aggregation choice.

Figure~\ref{fig:turn_surgery}B shows that 151 of 237 selected degraded trajectories (63.7\%) contain at least one positive surgery.
Among these 151 trajectories, 100 (66.2\%) contain multiple positive surgery positions, indicating that sensitivity is often distributed across several history locations rather than concentrated in a unique pivot turn.
This pattern is robust to an $\epsilon=.01$ practical tolerance for continuous-score tasks: 148 of 237 trajectories remain any-positive, and 95 of these 148 contain multiple positive positions.

For binary outcomes, 76 of 157 trajectories (48.4\%) contain at least one $0\rightarrow1$ fail-to-success surgery.
Thus, in nearly half of the selected binary trajectories, intervening on a single intermediate assistant turn can produce a counterfactual branch that reverses the final outcome.

\paragraph{Unedited-replay control.}
On 40 selected Qwen3-14B trajectories across five tasks, we compare newly generated Turn Surgery and unedited-replay continuations for 219 intervention branches.
Turn Surgery yields a trajectory-equal mean gain of $+.0847$ over unedited replay (95\% CI $[.0270,.1543]$).
Among these branches, 57 improve, 127 remain unchanged, and 35 are harmed.

\subsubsection{Turn-Level Selectivity and Position}

Broad trajectory-level reach coexists with strong turn-level selectivity.
Under trajectory-equal weighting, 31.5\% of tested positions produce a positive effect, 57.6\% produce no change, and 10.8\% produce a negative effect.
A positive surgery means that neutralizing the original turn improves the downstream score, whereas a negative surgery means that neutralization degrades it.
Thus, historical assistant turns are not interchangeable: many produce no change under this intervention, neutralizing some improves downstream performance, while neutralizing others harms it.

Figure~\ref{fig:turn_surgery}C shows how positive and negative effects vary across normalized assistant-turn positions.
Relative position has only a weak negative association with surgery effect ($\rho=-.139$).
Figure~\ref{fig:turn_surgery}D further shows that distance to the final user message is nearly unrelated to surgery effect ($\rho=-.005$).
History sensitivity therefore has positional structure but is not reducible to a simple recency rule, so proximity to the final request alone cannot identify the turns that matter.

\subsubsection{Stability of Selected Surgery Effects}

On the frozen 60-trajectory subset, 43 of 60 selected interventions (71.7\%) retain the same effect class across three runs, while 146 of 180 run pairs (81.1\%) agree.
These results show imperfect repeatability even at temperature $0$.

Taken together, RQ3 shows that history sensitivity is broad at the trajectory level but uneven and bidirectional across individual turns, without following a simple recency rule.
These findings argue against treating all assistant history uniformly.
Turn Surgery also provides turn-level sensitivity labels that could support future methods for selectively retaining, downweighting, or discarding prior assistant responses.

\subsection{RQ4: Internal Signatures Accompany Behavioral History Sensitivity}
\label{sec:results_rq4}

\begin{figure*}[t]
    \centering
    \includegraphics[width=\textwidth]{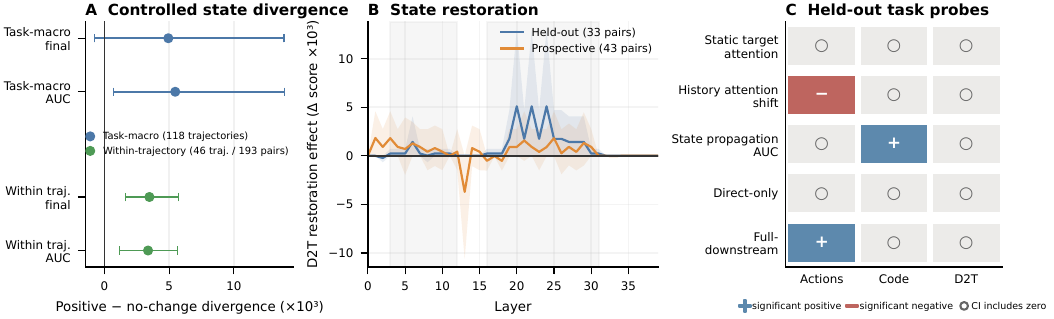}
    \caption{
    Internal changes accompany positive-effect history interventions, but their signatures are task dependent rather than universal.
    (A) Controlled state divergence between positive and no-change interventions.
    (B) State restoration profiles under held-out and prospective evaluations.
    (C) Held-out task-level outcomes across complementary probes for Actions, Code, and D2T.
    Symbols in panel C denote significant positive/negative effects or confidence intervals including zero.
    }
    \Description{
    Three panels present complementary analyses of internal model changes associated with history interventions.
    Panel A shows four point estimates with confidence intervals for positive-minus-no-change state divergence, comparing final-layer divergence and area under the curve (AUC) under task-macro and within-trajectory aggregation.
    All four point estimates are positive, with narrower confidence intervals for the within-trajectory comparisons.
    Panel B plots layer-wise D2T state restoration effects for 33 held-out pairs in blue and 43 prospective pairs in orange, with shaded uncertainty intervals.
    The held-out profile exhibits several positive peaks in intermediate-to-late layers, while the prospective profile shows smaller and less consistent positive effects.
    Panel C presents a matrix of five internal probes across Actions, Code, and D2T.
    Blue cells with plus signs indicate significant positive effects, red cells with minus signs indicate significant negative effects, and gray cells with circles indicate confidence intervals including zero.
    Actions exhibits a significant negative history attention shift and a significant positive full-downstream effect, while Code exhibits a significant positive state propagation AUC.
    All remaining task-probe combinations have confidence intervals including zero.
    }
    \label{fig:internal_signatures}
\end{figure*}

RQ4 examines the internal signatures of consequential history interventions.
Figure~\ref{fig:internal_signatures} summarizes downstream state divergence, restoration, and complementary probes in Qwen3-14B.

\subsubsection{Positive-Effect Interventions Produce Larger Downstream State Changes}

Controlled comparisons show larger downstream residual-state divergence for positive than for no-change surgeries.
The pooled AUC contrast is reversed at $-.00430$ because the groups differ in task composition, but task-balanced and within-trajectory comparisons recover the opposite direction (Figure~\ref{fig:internal_signatures}A).
The task-macro AUC contrast across Actions, Code, and D2T is $+.00547$ with a 95\% CI of $[.00069,.01390]$, and the within-trajectory comparison over 193 matched pairs from 46 trajectories yields $+.00337$ $[.00116,.00565]$.
The pattern remains under stricter positional controls: nearest-position matching yields $+.00271$ $[.00056,.00497]$, while matching with distance-to-final difference $\leq 1$ yields $+.00641$ $[.00344,.00996]$.
A separate prospective Data-to-Text analysis likewise shows greater divergence at the final downstream anchor, with a contrast of $+.00123$ $[.00079,.00167]$.

\subsubsection{State Restoration Provides a Bounded Causal Follow-Up}

Residual-state patching tests whether restoring a surgery-induced downstream state toward its original-history value undoes part of the surgery benefit (Figure~\ref{fig:internal_signatures}B).
For D2T, the held-out and prospective cohorts contain 33 and 43 pairs, respectively.
Across 40 layers, their restoration profiles correlate at $\rho=.456$, with 29 layers sharing the same sign.
However, prospective confidence intervals within the predefined regions cross zero.
Patching is therefore consistent with a D2T-specific directional link between downstream residual state and behavior, but does not establish mediation or a task-general restoration mechanism.

\subsubsection{No Single Internal Probe Is Consistent Across Tasks}

Figure~\ref{fig:internal_signatures}C shows that the held-out probes do not yield a single signature across tasks.
Static target-turn attention does not distinguish positive from no-change interventions.
Assistant-history attention redistribution is significant only in Actions, while state-propagation AUC is significantly positive only in Code; among the held-out knockout probes, only full-downstream intervention in Actions is significant.
In the prospective D2T analysis, attention reallocation is approximately null, whereas direct-only and full-downstream knockouts show sensitivity; the propagation advantage remains uncertain.
Continuation-margin shifts depend on the matching scheme and likewise do not provide a universal predictor.
Together, the internal signatures vary across tasks, probes, and evaluation cohorts rather than defining a task-general predictor.

\section{Discussion}
\label{sec:discussion}

Our results distinguish interaction robustness from one-shot capability and show that assistant-history effects vary across tasks and models.
Evaluation should therefore complement fully specified benchmarks with controlled multi-turn evaluation~\cite{laban2026llms,kwan-etal-2024-mt,DBLP:conf/acl/DeshpandeSMJHLK25}.
Unedited-replay controls further show positive average effects of history editing beyond ordinary regeneration on selected Qwen3-14B trajectories.
These findings highlight self-generated assistant history as part of the context shaping later decisions~\cite{huang2026llmsbenefitwords,zheng2026maigo,wan2026mitigating}.

Turn Surgery shows that individual assistant turns can have positive, negative, or negligible effects on downstream outcomes, without following a simple recency pattern.
Because uniform removal can discard helpful turns alongside harmful ones, these findings motivate selective history management~\cite{ma2026dover,bonagiri2026causalflow,shah2026causal}.

The open-weight analysis links behaviorally consequential history interventions to downstream representational differences.
State restoration and complementary internal probes reveal task-dependent effects rather than a single consistent mechanism across tasks.
More broadly, reliable interactive models must account for the context they help create and later consume~\cite{deng2024multi,yao2024tau,shinn2023reflexion}.

\subsection{Limitations}
\label{sec:limitations}

Our study has four main limitations.
First, replay interventions modify assistant history while preserving the observed user messages.
The resulting effects are conditional on these recorded messages and do not capture how users might adapt to different assistant responses.

Second, length-matched neutralization uses whitespace spans rather than exact model-token counts, and repeated neutral filler introduces a regular replacement pattern.
This control does not isolate all differences in token length, semantics, or context structure.
Repeatability analyses also cover selected interventions rather than all Turn Surgery branches.

Third, Turn Surgery identifies influential turns retrospectively through intervention and downstream evaluation.
Its effects are context-dependent and do not provide a validated online history selector.

Fourth, Turn Surgery is evaluated on selected degraded trajectories from six task families and five models in a single benchmark.
Their prevalence in the broader trajectory population and generalization to open-ended, longer-horizon agent interactions remain to be established.
The unedited-replay controls cover selected Qwen3-14B trajectories across five tasks, while internal analysis is limited to Qwen3-14B.

\section{Conclusion}

We study self-generated assistant history as a source of multi-turn reliability effects through cross-task evaluation, whole-history intervention, Turn Surgery, and open-weight internal analysis.
Multi-turn degradation exhibits a clear task $\times$ model structure, and assistant-history interventions change downstream outcomes, including relative to unedited replay on a selected Qwen3-14B subset.
Turn Surgery further shows that this sensitivity is broad across selected degraded trajectories but selective across individual history positions.
Open-weight analysis associates behaviorally consequential interventions with measurable downstream state differences while revealing task-dependent rather than universal internal signatures.
These findings motivate selective management of self-generated history in interactive systems.

\begin{acks}
This work was supported by the New Generation Artificial Intelligence--National Science and Technology Major Project (No.~2026ZD0127902) and the Fundamental and Interdisciplinary Disciplines Breakthrough Plan of the Ministry of Education of China (No.~JYB2025XDXM903).
\end{acks}

\clearpage

% =========================================================
% Ethical Considerations
% =========================================================
\section{Ethical Considerations}

This work studies multi-turn language-model behavior through controlled benchmark evaluation and counterfactual replay.
It does not involve deployment to real users, human-subject experimentation, or the collection of personal or sensitive user data.
Our interventions retrospectively modify self-generated assistant history within observed benchmark trajectories while preserving the user messages already present in those histories, with the goal of diagnosing downstream history effects.

The resulting causal claims are intervention-defined effects conditional on the observed user-message histories.
They do not estimate fully adaptive interactions in which a changed assistant response could also change subsequent user behavior, and benchmark history sensitivity should not be interpreted as direct evidence of safety, reliability, or failure rates in deployed systems.
Turn Surgery signals are likewise retrospective diagnostic quantities rather than validated online history-management decisions.
The internal analyses are additionally limited to the evaluated open-weight setting and should not be interpreted as complete explanations of model computation.
We therefore view these analyses as tools for studying interactive model behavior rather than as sufficient evidence for high-stakes deployment or automated decision-making.

% =========================================================
% References
% =========================================================

\bibliographystyle{ACM-Reference-Format}
\bibliography{reference}

% =========================================================
% Appendix
% =========================================================
%
% IMPORTANT:
% Appendices count toward the WSDM 2027 9-page limit.
%
% Uncomment only if needed.
%
% \appendix
% \input{sections/appendix}

\end{document}